\documentclass[runningheads]{llncs}
\usepackage{graphicx}
\usepackage{booktabs}
\usepackage{multirow}
\usepackage{hyperref}
\usepackage{xcolor}
\usepackage{algorithm}
\usepackage{algorithmicx}
\usepackage{amsmath}
\usepackage{amssymb}
\usepackage{nicefrac}

\DeclareMathOperator*{\softmax}{softmax}

\usepackage[acronym]{glossaries}
\newacronym{kg}{KG}{Knowledge Graph}
\newacronym{ea}{EA}{Entity Alignment}
\newacronym{rdgcn}{RDGCN}{Relation-aware Dual-Graph Convolutional Network}
\newacronym{dgmc}{DGMC}{Deep Graph Matching Consensus}
\newacronym{gcnalign}{GCN-Align*}{GCN-Align*}
\newacronym{gnn}{GNN}{Graph Neural Network}
\newacronym{sota}{SotA}{state-of-the-art}

\glsdisablehyper

\newcommand{\tablesize}[0]{
\scriptsize
}

\begin{document}
\title{
A Critical Assessment of State-of-the-Art in Entity Alignment
}
\author{
Max Berrendorf\inst{1}\orcidID{0000-0001-9724-4009}
\and
Ludwig Wacker\inst{1}\and
Evgeniy Faerman\inst{1}}
\authorrunning{M. Berrendorf et al.}
\institute{
Ludwig-Maximilians-Universität München, Munich, Germany\\ \email{\{berrendorf,faerman\}@dbs.ifi.lmu.de}, \email{l.wacker@campus.lmu.de}
}
\maketitle              \begin{abstract}
In this work, we perform an extensive investigation of two \gls{sota} methods for the task of \acrlong{ea} in \acrlong{kg}s.
Therefore, we first carefully examine the benchmarking process and identify several shortcomings, making the results reported in the original works not always comparable. 
Furthermore, we suspect that it is a common practice in the community to make the hyperparameter optimization directly on a test set, reducing the informative value of reported performance.
Thus, we select a representative sample of benchmarking datasets and describe their properties.
We also examine different initializations for entity representations since they are a decisive factor for model performance. 
Furthermore, we use a shared train/validation/test split for an appropriate evaluation setting to evaluate all methods on all datasets.
In our evaluation, we make several interesting findings.
While we observe that most of the time \gls{sota} approaches perform better than baselines, they have difficulties when the dataset contains noise, which is the case in most real-life applications. Moreover, in our ablation study,  we find out that often different features of \gls{sota} method are crucial for good performance than previously assumed.
The code is available at \url{https://github.com/mberr/ea-sota-comparison}.
\end{abstract}

\section{Introduction}
The quality of information retrieval crucially depends on the accessible storage of information.
\glspl{kg} often serve as such data structure~\cite{Dietz2019}.
Moreover, to satisfy diverse information needs, a combination of multiple data sources is often inevitable.
\gls{ea}~\cite{DBLP:conf/ecir/BerrendorfFMT020} is the discipline of aligning entities from different \glspl{kg}.
Once aligned, these entities facilitate information transfer between knowledge bases, or even fusing multiple \glspl{kg} to a single knowledge base.

In this work, our goal is to analyze a \gls{sota} approach for the task of \gls{ea} and identify which factors are essential for its performance. 
Although papers often use the same dataset in the evaluation and report the same evaluation metrics, the selection of \gls{sota} is not a trivial task:
as we found out in our analysis, the usage of different types of external information for the initialization or train/test splits of different sizes\footnote{Commonly used evaluation metrics in \acrshort{ea} automatically become better with a smaller size of test set~\cite{DBLP:journals/corr/abs-2002-06914}.} makes the results in different works incomparable. 
Therefore, while still guided by the reported evaluation metrics, we identified these common factors among strongly performing methods in multiple works:
\begin{itemize}
    \item They are based on \glspl{gnn}. \glspl{gnn} build the basis of the most recent works~\cite{DBLP:conf/emnlp/WangLLZ18,DBLP:conf/emnlp/LiCHSLC19,DBLP:conf/emnlp/ShiX19,DBLP:conf/emnlp/YangZSLLS19,DBLP:conf/ijcai/Ye0FZW19,DBLP:conf/acl/CaoLLLLC19,DBLP:conf/ijcai/ZhangSHCGQ19,DBLP:conf/ijcai/ZhuZ0TG19,DBLP:conf/wsdm/MaoWXLW20,DBLP:conf/aaai/SunW0CDZQ20,DBLP:conf/iclr/FeyL0MK20,DBLP:conf/ijcai/WuLF0Y019,DBLP:conf/acl/XuWYFSWY19,DBLP:conf/emnlp/WuLFWZ19,DBLP:journals/corr/abs-2010-04348}.
    \item They utilize entity names in the model. Supported by recent advances in word embeddings, these attributes provide distinctive features.
    \item They consider different types of relations existing in \glspl{kg}. Most \glspl{gnn} ignore different relationship types and aggregate them in the preprocessing step.
\end{itemize}
Given these criteria, we selected \acrfull{rdgcn}~\cite{DBLP:conf/ijcai/WuLF0Y019}, as it also has demonstrated impressive performance in recent benchmarking studies~\cite{DBLP:journals/pvldb/SunZHWCAL20,tkde/Zhao2020}.
Additionally, we include the recently published \acrfull{dgmc}~\cite{DBLP:conf/iclr/FeyL0MK20} method in our analysis for two reasons:
the studies mentioned above did not include it, and the authors reported surprisingly good performance, considering that this method does not make use of relation type information.

We start our study by reviewing the used datasets and discussing the initializations based on entity names.
Although both methods utilize entity names, the actual usage differs.
For comparison, we thus evaluate both methods on all datasets with all available initializations.
We also report the zero-shot performance, i.e., when only using initial representations alone, as well as a simple \gls{gnn} model baseline.
Furthermore, we address the problem of hyperparameter optimization. 
Related works often do not discuss how they chose hyperparameters and, e.g., rarely report validation splits. So far, this problem was not addressed in the community. In the recent comprehensive survey~\cite{DBLP:journals/pvldb/SunZHWCAL20}, the authors use cross-validation for the estimation of the test performance. The models are either evaluated with hyperparameters recommended for other datasets or selected by not reported procedure.
Also, in the published code of the investigated approaches, we could not find any trace of train-validation splits, raising questions about reproducibility and fairness of their comparisons.
We thus create a shared split with a test, train, and validation part and extensively tune the model's hyperparameters for each of the dataset/initialization combinations to ensure that they are sufficiently optimized.
Finally, we provide an ablation study for many of the parameters of a \gls{sota} approach (\acrshort{rdgcn}), giving insight into the individual components' contributions to the final performance.

\section{Datasets \& Initialization}
\label{sec:datasets_inits}
\begin{table}
\centering
\caption{
Summary of the used \gls{ea} datasets.
We denote the entity set as $\mathcal{E}$, the relation set as $\mathcal{R}$, the triple set as $\mathcal{T}$, the aligned entities as $\mathcal{A}$ and the exclusive entities as $\mathcal{X}$.
}
\label{tab:datasets}
\tablesize{}
\begin{tabular*}{\textwidth}{l*{2}{@{\extracolsep{\fill} }l}*{5}{@{\extracolsep{\fill} }r}}
\toprule
dataset & subset & graph& $|\mathcal{E}|$ & $|\mathcal{R}|$ & $|\mathcal{T}|$ & $|\mathcal{A}|$ & $|\mathcal{X}|$ \\
\midrule
DBP15k & zh-en & zh &          19,388 &           1,701 &          70,414 &          15,000 &           4,388 \\
       &     & en &          19,572 &           1,323 &          95,142 &          15,000 &           4,572 \\
       & ja-en & ja &          19,814 &           1,299 &          77,214 &          15,000 &           4,814 \\
       &     & en &          19,780 &           1,153 &          93,484 &          15,000 &           4,780 \\
       & fr-en & fr &          19,661 &             903 &         105,998 &          15,000 &           4,661 \\
       &     & en &          19,993 &           1,208 &         115,722 &          15,000 &           4,993 \\
\midrule
WK3l15k & en-de & en &          15,126 &           1,841 &         209,041 &           9,783 &           5,343 \\
       &     & de &          14,603 &             596 &         144,244 &          10,021 &           4,582 \\
       & en-fr & en &          15,169 &           2,228 &         203,356 &           7,375 &           7,794 \\
       &     & fr &          15,393 &           2,422 &         169,329 &           7,284 &           8,109 \\
\midrule
OpenEA & en-de & en &          15,000 &             169 &          84,867 &          15,000 &               0 \\
       &     & de &          15,000 &              96 &          92,632 &          15,000 &               0 \\
       & en-fr & en &          15,000 &             193 &          96,318 &          15,000 &               0 \\
       &     & fr &          15,000 &             166 &          80,112 &          15,000 &               0 \\
       & d-y & d &          15,000 &              72 &          68,063 &          15,000 &               0 \\
       &     & y &          15,000 &              21 &          60,970 &          15,000 &               0 \\
       & d-w & d &          15,000 &             167 &          73,983 &          15,000 &               0 \\
       &     & w &          15,000 &             121 &          83,365 &          15,000 &               0 \\
\bottomrule
\end{tabular*} \end{table}
Table~\ref{tab:datasets} provides a summary of a representative sample of datasets used for benchmarking of \gls{ea} approaches.
In the following, we first discuss each dataset's properties and, in the second part, the initialization of entity name attributes.

\subsection{Datasets}
\paragraph{DBP15k}
The DBP15k dataset is the most popular dataset for the evaluation of \gls{ea} approaches.
It has three subsets, all of which base upon DBpedia. 
Each subset comprises a pair of graphs from different languages. 
As noted by \cite{DBLP:conf/ecir/BerrendorfFMT020}, there exist multiple variations of the dataset, sharing the same entity alignment but differing in the number of exclusive entities in each graph.
The alignments in the datasets are always 1:1 alignments, and due to the construction method for the datasets, exclusive entities do not have relations between them, but only to shared entities. 
Exclusive entities complicate the matching process, and in real-life applications, they are not easy to identify.
Therefore, we believe that this dataset describes a realistic use-case only to a certain extent.
We found another different variant of DBP15k as part of the PyTorch Geometric repository\footnote{\url{https://github.com/rusty1s/pytorch\_geometric/blob/d42a690fba68005f5738008a04f375ffd39bbb76/torch\_geometric/datasets/dbp15k.py}}, having a different set of aligned entities.
This is likely due to extraction of alignments from data provided by~\cite{DBLP:conf/acl/XuWYFSWY19} via Google Drive\footnote{\url{https://drive.google.com/open?id=1dYJtj1\_J4nYJdrDY95ucGLCuZXDXI7PL}} as described in their GitHub repository.\footnote{\url{https://github.com/syxu828/Crosslingula-KG-Matching/blob/56710f8131ae072f00de97eb737315e4ac9510f2/README.md\#how-to-run-the-codes}}
As a result, the evaluation results published in \cite{DBLP:conf/iclr/FeyL0MK20} are not directly comparable to other published results. 
In our experiments, we use the (smaller) JAPE variant with approximately 19-20k entities in each graph since it is the predominantly used variant.

\paragraph{OpenEA}
The OpenEA datasets published by~\cite{DBLP:journals/pvldb/SunZHWCAL20} comprise graph pairs from DBPedia, YAGO, and Wikidata obtained by iterative degree-based sampling to match the degree distribution between the source \gls{kg} and the extracted subset.
The alignments are exclusively 1:1 matchings, and there are no exclusive entities, i.e., every entity occurs in both graphs.
We believe that this is a relatively unrealistic scenario.
In our experiments, we use all graph pairs with 15k entities (\texttt{15K}) in the dense variant (\texttt{V2}), i.e., \texttt{en-de-15k-v2}, \texttt{en-fr-15k-v2}, \texttt{d-y-15k-v2}, \texttt{d-w-15k-v2}.

\paragraph{WK3l15k}
The Wk3l datasets are multi-lingual \gls{kg} pairs extracted from Wikipedia.
As in~\cite{DBLP:conf/ecir/BerrendorfFMT020}, we extract additional entity alignments from the triple alignments.
The graphs contain additional exclusive entities, and there are m:n matchings.
We only use the 15k variants, where each graph has approximately 15k entities.
There are two graph pairs, \texttt{en-de} and \texttt{en-fr}.
Moreover, the alignments in the dataset are relatively noisy:
for example, \texttt{en-de} contains besides valid alignments such as (``trieste", ``triest"), or (``frederick i, holy roman emperor", ``friedrich i. (hrr)"), also ambiguous ones such as (``1", ``1. fc saarbrücken"), (``1", ``1. fc schweinfurt 05"), and errors such as (``1", ``157"), and (``101", ``100").
While the noise aggravates alignment, it also reflects a realistic setting.

\subsection{Label-Based Initializations}
\label{sec:init}

\paragraph{Prepared translations (DBP15k)}
For DBP15k, we investigate label-based initializations based on prepared translations to English from \cite{DBLP:conf/ijcai/WuLF0Y019} and \cite{DBLP:conf/iclr/FeyL0MK20} (which, in turn, originate from \cite{DBLP:conf/acl/XuWYFSWY19}).
Afterwards, they use Glove~\cite{DBLP:conf/emnlp/PenningtonSM14} embeddings to obtain an entity representation.
While \cite{DBLP:conf/ijcai/WuLF0Y019} only provides the final entity representation vectors without further describing the aggregation, \cite{DBLP:conf/iclr/FeyL0MK20} splits the label into words (by white-space) and uses the sum over the words' embeddings as entity representation.
\cite{DBLP:conf/ijcai/WuLF0Y019} additionally normalizes the norm of the representations to unit length.

\paragraph{Prepared \acrshort{rdgcn} Embeddings (OpenEA)}
\begin{table}[t]
    \centering
    \caption{
    The statistics about label-based initialization in the OpenEA codebase:
    \emph{attribute} denotes initialization via attribute values for a predefined set of ``name attributes".
    \emph{id} denotes initialization with the last part of the entity URI.
    For \texttt{d-y} this basically leaks ground truth, whereas, for Wikidata, the URI contains only a numeric identifier, thus rendering the initialization ``label" useless.
    }
    \label{tab:openea:labels}
    \tablesize{}
    \begin{tabular*}{\textwidth}{*{2}{l@{\extracolsep{\fill}}}*{3}{@{\extracolsep{\fill}}r}}
\toprule
subset & side & via attribute & via id &  via id (\%) \\
\midrule
\multirow{2}{*}{d-w}
    & d &      0 & 15,000 & 100.00\% \\
    & w &  8,391 &  7,301 &  48.67\% \\
\multirow{2}{*}{d-y}
    & d &  2,883 & 12,122 &  80.81\% \\
    & y & 15,000 &      0 &   0.00\% \\
\bottomrule
\end{tabular*}
 \end{table}
OpenEA~\cite{DBLP:journals/pvldb/SunZHWCAL20} benchmarks a large variety of contemporary entity alignment methods in a unified setting, also including \acrshort{rdgcn}~\cite{DBLP:conf/ijcai/WuLF0Y019}.
Since the graphs DBPedia and YAGO collect data from similar sources, the labels are usually equal.
For those graph pairs, the authors propose to delete the labels.
However, \acrshort{rdgcn} requires a label based initialization.
Thus, the authors obtain labels via attribute triples of a pre-defined set of ``name-attributes"\footnote{\url{https://github.com/nju-websoft/OpenEA/tree/2a6e0b03ec8cdcad4920704d1c38547a3ad72abe}}:
{\ttfamily skos:prefLabel, http://dbpedia.org/ontology/birth\-Name} for DBPedia-YAGO, and {\ttfamily http://www.wiki\-data.org/en\-ti\-ty/P373, http:\-//www.wiki\-data.org/entity/P1476} for DBPedia-Wikidata.

However, when investigating the published code, we noticed that if the label is not found via attribute, the last part of the entity URI is used instead.
For DBPedia/YAGO, this effectively leaks ground truth since they share the same label.
For DBPedia/Wikidata, this results in useless labels for the Wikidata side since their labels are the Wikidata IDs, e.g., \texttt{Q3391163}.
Table~\ref{tab:openea:labels} summarizes the frequency of both cases.
For \texttt{d-w}, DPBedia entities always use the ground truth label. 
For 49\% of the Wikidata entities, useless labels are used for initialization.
For \texttt{d-y}, YAGO entity representations are always initialized via an attribute triple.
For DBPedia, in 81\% of all cases, the ground truth label is used.
We store these initial entity representations produced by the OpenEA codebase into a file and refer in the following to them as \emph{Sun} initialization (since they are taken from the implementation of \cite{DBLP:journals/pvldb/SunZHWCAL20}).

\paragraph{Multi-lingual BERT (WK3l15k)}
Since we did not find related work with entity embedding initialization from labels on WK3l15k, we generated those using a pre-trained multi-lingual BERT model~\cite{DBLP:conf/naacl/DevlinCLT19}, \texttt{BERT-Base, Multilingual Cased}\footnote{\url{https://github.com/google-research/bert/blob/cc7051dc592802f501e8a6f71f8fb3cf9de95dc9/multilingual.md}}.
Following~\cite{DBLP:conf/naacl/DevlinCLT19}, we use the sum of the last four layers as token representation since it has comparable performance to the concatenation at a quarter of its size.
To summarize the token representations of a single entity label, we explore sum, mean, and max aggregation as hyperparameters.

\section{Methods}
\label{sec:methods}
We evaluate two \gls{sota} \gls{ea} methods, \acrshort{rdgcn}~\cite{DBLP:conf/ijcai/WuLF0Y019} which we reimplemented and \acrshort{dgmc}~\cite{DBLP:conf/iclr/FeyL0MK20} for which we used the original method implementation with adapted evaluation.
In the following, we revisit their architectures and highlight differences between the architecture described in the paper and what we found in the published code.

Similarly to all \gls{gnn}-based approaches, both models employ a Siamese architecture. 
Therefore, the same model with the same weights is applied to both graphs yielding representations of entities from both \glspl{kg}. 
Given these entity representations, the \gls{ea} approaches compute an affinity matrix that describes the similarity of entity representations from both graphs.
Since the main difference between methods is the \gls{gnn} model in the Siamese architecture, for brevity we only describe how it is applied on a single \gls{kg} $\mathcal{G} = (\mathcal{E}, \mathcal{R}, \mathcal{T})$.

\subsection{\acrfull{rdgcn}}

\subsubsection{Architecture}
The \acrshort{rdgcn}~\cite{DBLP:conf/ijcai/WuLF0Y019} model comprises two parts performing message-passing processes applied sequentially.
The message passing process performed by the first part can be seen as \emph{relation-aware}.
The model tries to learn the importance of relations and weights the messages from the entities connected by these relations correspondingly. 
The message passing performed by the second component utilizes a simple adjacency matrix indicating the existence of any relations between entities, which we call \emph{standard message passing}.
Both components employ a form of skip connections: (weighted) residual connections~\cite{DBLP:conf/cvpr/HeZRS16} in the first part and highway layers~\cite{DBLP:journals/corr/SrivastavaGS15} in the second part.

\paragraph{Relation-Aware Message Passing}
The entity embeddings from the first component 
are computed by several \emph{interaction rounds} comprising four steps
\begin{eqnarray}
    \mathbf{X_c} &=& RC(\mathbf{X_e}), \mathbf{X_c} \in \mathbb{R}^{|\mathcal{R}| \times 2d} \label{eq:rdgcn:interaction:context} \\ \mathbf{X_r} &=& DA(\mathbf{X_r}, \mathbf{X_c}),  \mathbf{X_r} \in \mathbb{R}^{|\mathcal{R}| \times 2d} \label{eq:rdgcn:interaction:dual} \\ \mathbf{X_e} &=& PA(\mathbf{X_e}, \mathbf{X_r}) \label{eq:rdgcn:interaction:primal} \\ \mathbf{X_e} &=& \mathbf{X_e^{0}} + \beta_i \cdot \mathbf{X_e} \label{eq:rdgcn:interaction:skip}\end{eqnarray}
The first step, in \eqref{eq:rdgcn:interaction:context}, obtains a \emph{relation context} (RC) $\mathbf{X_c}$ from the entity representations.
For relation $r \in \mathcal{R}$, we extract its relation context as a concatenation of the mean entity representations for the head and the tail entities.
By denoting the set of head and tail entities for relation $r$ with $H_r$ and $T_r$, we can thus express its computation as
$
    (\mathbf{X_c})_i = \left[
    \nicefrac{1}{|H_i|} \sum_{j \in H_i} (\mathbf{X_e})_j
    ~\|~
    \nicefrac{1}{|T_i|} \sum_{j \in T_i} (\mathbf{X_e})_j
    \right]
$ where $\|$ denotes the concatenation operation.
An entity occurring multiple times as the head is weighted equally to an entity occurring only once.

The second step, in \eqref{eq:rdgcn:interaction:dual}, is the \emph{dual graph attention} (DA).
The attention scores on the dual graph $\alpha_{ij}^D$ are computed by dot product attention with leaky ReLU activation:
$
    \alpha_{ij}^D = J_{ij} \cdot LeakyReLU(\mathbf{W_L} (\mathbf{X_c})_i + \mathbf{W_R} (\mathbf{X_c})_j).
$
Notice that $
    \mathbf{W_L} (\mathbf{X_c})_i + \mathbf{W_R} (\mathbf{X_c})_j = (\mathbf{W_L} \| \mathbf{W_R})^T ((\mathbf{X_c})_i \| (\mathbf{X_c})_j)
$, where $\|$ denotes the concatenation operation.
In the published code, we further found a weight sharing mechanism for $\mathbf{W_L}$ and $\mathbf{W_R}$ implemented,
decomposing the projection weight matrices as $\mathbf{W_L} = \mathbf{W_L'} \mathbf{W_C}$ and $\mathbf{W_R} = \mathbf{W_R'} \mathbf{W_C}$ with $\mathbf{W_L'}, \mathbf{W_R'} \in \mathbb{R}^{1 \times h}, \mathbf{W_C} \in \mathbb{R}^{h \times 2d}$ being trainable parameters, and $\mathbf{W_C}$ shared between both projections.
$J_{ij}$ denotes a fixed triple-based relation similarity score computed as the sum of the Jaccard similarities of the head and tail entity set for relation $r_i$ and $r_j$:
$
    J_{ij} := \nicefrac{|H_i \cap H_j|}{|H_i \cup H_j|} + \nicefrac{|T_i \cap T_j|}{|T_i \cup T_j|}.
$
The softmax is then computed only over those relations, where $J_{ij} > 0$, i.e., pairs sharing at least one head or tail entity.
In the implementation, this is implemented as dense attention with masking, i.e. setting $\alpha_{ij}^D = -\infty$ (or a very small value) for $J_{ij} = 0$.
While this increases the required memory consumption to $\mathcal{O}(|\mathcal{R}|^2)$, the number of relations is usually small compared to the number of entities, cf. Table~\ref{tab:datasets}, and thus this poses no serious computational problem.
With $\tilde{\alpha}_{ij}^D$ denoting the softmax output, the new relation representation finally is
$
    (\mathbf{X_r})_i = ReLU\left(\sum_{j} \tilde{\alpha}_{ij}^D (\mathbf{X_r})_j\right).
$

In the third step, in \eqref{eq:rdgcn:interaction:primal}, the entity representations are updated.
To this end, a relation-specific scalar score is computed as
$
    \alpha_i^r = LeakyReLU(\mathbf{W} \mathbf{X_r} + b)
$
with trainable parameters $W$ and $b$. 
Based upon the relation-specific scores, an attention score between two entities $e_i, e_j$ with at least one relation between them is given as
$
    \alpha_{ij}^P = \sum_{r \in \mathcal{T}_{ij}} \alpha_i^r.
$
These scores are normalized with a sparse softmax over all $\{j \mid \exists r \in \mathcal{R}: (e_i, r, e_j) \in \mathcal{T}\}$:
$
    \tilde{\alpha}_{ij}^P = \softmax_{j'}(\alpha_{ij'}^P)_j.
$
The final output of the primal attention is
$
    (\mathbf{X_e})_j = ReLU(\sum_i \tilde{\alpha}_{ij} (\mathbf{X_e})_j).
$

The fourth step, in \eqref{eq:rdgcn:interaction:skip}, applies a skip connection from the initial representations to the current entity representation.
The weight $\beta_i$ is pre-defined ($\beta_1 = 0.1$, $\beta_2 = 0.3$) and not trained.

\paragraph{Standard Message Passing}
The second part of the \acrshort{rdgcn} consists of a sequence of GCN layers with highway layers.
Each layer computes
\begin{eqnarray}
    \mathbf{X_e'} &=& ReLU(\mathbf{A} \mathbf{X_e} \mathbf{W}) \\
    \beta &=& \sigma(\mathbf{W_g} \mathbf{X_e} + b_g) \\
    \mathbf{X_e} &=& \beta \cdot \mathbf{X_e'} + (1 - \beta) \cdot \mathbf{X_e}
\end{eqnarray}
$\mathbf{A} \in \mathbb{R}^{|\mathcal{E}^L| \times |\mathcal{E}^L|}$ denotes the adjacency matrix of the primal graph.
It is constructed by first creating an undirected, unweighted adjacency matrix where there is a connection between $e_i, e_j \in \mathcal{E}^L$ if there exists at least one triple $(e_i, r, e_j) \in \mathcal{T}^L$ for some relation $r \in \mathcal{R}^L$.
Next, self-loops $(e, e)$ are added for every entity $e \in \mathcal{E}^L$.
Finally, the matrix is normalized by setting $\mathbf{A} = \mathbf{D}^{-1/2}\mathbf{A}\mathbf{D}^{-1/2}$ with $\mathbf{D}$ denoting the diagonal matrix of node degrees.
When investigating the published code, we further found out that the weight matrix $\mathbf{W}$ is constrained to be a diagonal matrix and initialized as an identity matrix.

\subsubsection{Training}
Let $\mathbf{x}_i^L$ denote the final entity representation for $e_i^L \in \mathcal{E}^L$ and anologously $\mathbf{x}_j^R$ for $e_j^R \in \mathcal{E}^R$.
\acrshort{rdgcn} is trained with a margin-based loss formulation.
It adopts a hard negative mining strategy, i.e., the set of negative examples for one pair is the top $k$ most similar entities of one of the entities according to the similarity measure used for scoring. The negative $l_1$ distance is used as similarity, the margin is 1, $k=10$, and the negative examples are updated every 10 epochs.

\subsection{\acrfull{dgmc}}
\acrshort{dgmc}~\cite{DBLP:conf/iclr/FeyL0MK20} also comprises two parts, which we name \emph{enrichment} and \emph{correspondence refinement}.
The enrichment part is a sequence of \gls{gnn} layers enriching the entity representations with information from their neighborhood.
Each layer computes
$
\phi(\mathbf{X}) = ReLU(norm(\mathbf{A})\mathbf{X}\mathbf{W_1} + norm(\mathbf{A}^T)\mathbf{X}\mathbf{W_2} + \mathbf{X}\mathbf{W_3})
$,
where $\mathbf{A} \in \mathbb{R}^{|\mathcal{E}^L| \times |\mathcal{E}^L|}$ denotes the symmetrically normalized adjacency matrix (as for second part of \acrshort{rdgcn}), $norm$ the row-wise normalization operation, $\mathbf{X} \in \mathbb{R}^{\mathcal{E}^L \times d_{in}}$ the layer's input, and $\mathbf{W_1}, \mathbf{W_2}, \mathbf{W_3} \in \mathbb{R}^{d_{in} \times d_{out}}$ trainable parameters of the layer.
An optional batch normalization and dropout follow this layer.
For the enrichment phase's final output, all individual layers' outputs are concatenated before a learned final linear projection layer reduces the dimension to $d_{out}$.

The second phase, the \emph{correspondence refinement}, first calculates the $k=10$ most likely matches in the other graph for each entity as a sparse correspondence matrix $\mathbf{S} \in \mathbb{R}^{|\mathcal{E}^L| \times |\mathcal{E}^R|}$, normalized using softmax.
Next, it generates random vectors for each entity $\mathbf{R} \in \mathbb{R}^{|\mathcal{E}^L| \times d_{rnd}}$ and sends these vectors to the probable matches via the softmax normalized sparse correspondence matrix, $\mathbf{S}^T\mathbf{R} \in \mathbb{R}^{|\mathcal{E}^R| \times d_{rnd}}$.
A \gls{gnn} layer $\psi$ as in phase one distributes these vectors in the neighborhood of the nodes: $\mathbf{Y}^R = \psi(\mathbf{S}^T\mathbf{R})$.
A two-layer MLP predicts an update for the correspondence matrix, given the difference between the representations $\mathbf{Y}^L$ and $\mathbf{Y}^R$.
This procedure is repeated for a fixed number of refinement steps $L=10$.

\section{Experiments}
\label{sec:experiments}
\paragraph{Experimental Setup}
\begin{table}[t]
\centering
\caption{
Investigated hyperparameters for all methods. * denotes that these parameters share the same value range but were tuned independently.
}
\label{tab:hparam}
\tablesize{}
\begin{tabular*}{\textwidth}{lp{.5\textwidth}}
    \toprule
    \multicolumn{2}{c}{\textbf{Common}} \\
    parameter & choices \\
    \midrule
    optimizer & Adam \\
    similarity & \{cos, dot, l1 (bound inverse), l1 (negative), l2 (bound inverse), l2 (negative)\} \\
    \toprule
    \multicolumn{2}{c}{\textbf{\acrshort{rdgcn}}} \\
    parameter & choices \\
    \midrule
    (entity embedding) normalization & \{always-l2, initial-l2, never\} \\
    (number of) GCN layers & \{0, 1, 2, 3\} \\
    (number of) interaction layers & \{0, 1, 2, 3\} \\
    interaction weights & $\{0.1, 0.2, \ldots, 0.6\}$ \\
    trainable embeddings & \{False, True\} \\
hard negatives & \{no, yes\} \\
    learning rate & $[10^{-4}, 10^{-1}]$\\
    \midrule
    \multicolumn{2}{c}{\textbf{\acrshort{dgmc}}} \\
    parameter & choices \\
    \midrule
    $\psi_1$ / $\psi_2$ dimension* & $[32, 64, \ldots, 1024]$ \\
    $\psi_1$ / $\psi_2$  (number of) GCN layers* & \{1, 2, 3, 4, 5\} \\
    $\psi_1$ / $\psi_2$  batch normalization* & \{False, True\} \\
    $\psi_1$  / $\psi_2$ layer concatenation* & \{False, True\} \\
    $\psi_1$ dropout & $[0.00, 0.05, \ldots, 1.0]$ \\
    $\psi_2$ dropout & 0.0 \\
    trainable embeddings & False \\
    (entity embedding) normalization & \{never, always-l1, always-l2\} \\
    learning rate & $[10^{-3}, 10^{-1}]$ \\
    \midrule
    \multicolumn{2}{c}{\textbf{\acrshort{gcnalign}}} \\
    parameter & choices \\
    \midrule
    model output dimension & $[32, 64, \ldots, (embedding dimension)]$ \\
    (number of) GCN layers & \{1, 2, 3\} \\
    batch normalization & \{False, True\} \\
    layer concatenation & \{False, True\} \\
    final linear projection & \{False, True\} \\
    dropout & $\{0.0, 0.1, \ldots, 0.5\}$ \\
    trainable embeddings &  \{False, True\} \\
    (entity embedding) normalization & \{never, always-l1, always-l2\} \\
    (weight) sharing horizontal & \{False, True\} \\
    learning rate & $[10^{-3}, 10^{-1}]$ \\
    \midrule
    \bottomrule
\end{tabular*} \end{table}
For the general evaluation setting and description of metrics, we refer to~\cite{DBLP:journals/corr/abs-2002-06914}.
Here, we primarily use Hits@1 (H@1), which measures the correct entity's relative frequency of being ranked in the first position.
When investigating the published code of both, \acrshort{rdgcn}~\cite{DBLP:conf/ijcai/WuLF0Y019}\footnote{\url{https://github.com/StephanieWyt/RDGCN}} and \acrshort{dgmc}~\cite{DBLP:conf/iclr/FeyL0MK20}\footnote{\url{https://github.com/rusty1s/deep-graph-matching-consensus/}}, we did not find any code for tuning the parameters, nor a train-validation split.
Also, the papers themselves do not mention a train-validation split.
Thus, it is unclear how they choose the hyperparameters without a test-leakage by directly optimizing the test set's performance.
We thus decided to create a shared test-train-validation split used by all our experiments to enable a fair comparison.
Since \acrshort{dgmc} already uses PyTorch, we could use their published code and extend it with HPO code.
\acrshort{rdgcn} was re-implemented in PyTorch in our codebase.
We use the official train-test split for all datasets, which reserves 70\% of the alignments for testing.
We split the remaining part into 80\% train alignments and 20\% validation alignments.

We continued by tuning numerous model parameters (cf. Table~\ref{tab:hparam}) of all models on each of the datasets in Table~\ref{tab:datasets} and each of the available initializations described in Section~\ref{sec:init} to obtain sufficiently well-tuned configurations.
We used random search due to its higher sample efficiency than grid search~\cite{DBLP:journals/jmlr/BergstraB12}.
We additionally evaluate a baseline, which uses the \gls{gnn} variant from \acrshort{dgmc} without the neighborhood consensus refinement, coined \emph{\acrshort{gcnalign}} due to its close correspondence to \cite{DBLP:conf/emnlp/WangLLZ18}, and also evaluate the zero-shot performance of the initial node features.

For each tested configuration, we perform early stopping on validation H@1, i.e., select the epoch according to the best validation H@1.
Across all tested configurations for a model-dataset-initialization combination, we then choose the best configuration according to validation H@1 and report the test performance in Table~\ref{tab:res:all}. 
We do not report performance for training on train+validation with the final configuration due to space restrictions.
We decided to report performance when trained only on the train set to ensure that other works have performance numbers for comparison when tuning their own models.

\subsection{Results}
\begin{table}
    \centering
    \caption{
    Results in terms of H@1 for all investigated combinations of datasets, models, and initializations.
    Each cell represents the \emph{test} performance of the best configuration of hyperparameters chosen according to \emph{validation} performance.
    }
    \label{tab:res:all}
    \tablesize{}
    DBP15k (JAPE) \\
    \begin{tabular*}{\textwidth}{l*{6}{@{\extracolsep{\fill}}r}}
\toprule
init & \multicolumn{3}{c}{Wu~\cite{DBLP:conf/emnlp/WuLFWZ19}} & \multicolumn{3}{c}{Xu~\cite{DBLP:conf/acl/XuWYFSWY19}} \\
subset &           fr-en &           ja-en &           zh-en &           fr-en &           ja-en &           zh-en \\
\midrule
Zero Shot           &           79.47 &           63.48 &           56.07 &           83.70 &           65.64 &           59.40 \\
\acrshort{gcnalign} &           81.81 &           67.45 &           57.94 &           86.74 &           67.65 &           60.32 \\
\acrshort{rdgcn}    &           86.91 &  \textbf{72.90} &           66.44 &           86.82 &           74.35 &  \textbf{69.54} \\
\acrshort{dgmc}     &  \textbf{89.35} &           72.17 &  \textbf{69.98} &  \textbf{90.12} &  \textbf{76.60} &           68.76 \\
\bottomrule
\end{tabular*}
     \\[1em]
    OpenEA \\
    \begin{tabular*}{\textwidth}{l*{4}{@{\extracolsep{\fill}}r}}
\toprule
init & \multicolumn{4}{c}{Sun~\cite{DBLP:journals/pvldb/SunZHWCAL20}} \\
subset &             d-w &             d-y &           en-de &           en-fr \\
\midrule
Zero Shot           &           46.53 &           81.90 &           75.99 &           79.90 \\
\acrshort{gcnalign} &           45.76 &           84.65 &           85.34 &           89.41 \\
\acrshort{rdgcn}    &  \textbf{64.28} &  \textbf{98.41} &           80.03 &  \textbf{91.52} \\
\acrshort{dgmc}     &           51.29 &           88.60 &  \textbf{88.10} &           89.40 \\
\bottomrule
\end{tabular*}
     \\[1em]
    WK3l15k \\
    \begin{tabular*}{\textwidth}{l*{2}{@{\extracolsep{\fill}}r}}
\toprule
init & \multicolumn{2}{c}{BERT} \\
subset &           en-de &           en-fr \\
\midrule
Zero Shot           &           85.55 &           77.27 \\
\acrshort{gcnalign} &           85.92 &  \textbf{78.22} \\
\acrshort{rdgcn}    &  \textbf{86.76} &           78.05 \\
\acrshort{dgmc}     &           84.08 &           73.92 \\
\bottomrule
\end{tabular*}
 \end{table}
Table~\ref{tab:res:all} presents the overall results.
We can observe several points.

\paragraph{Zero-Shot Performance}
Generally, there is an impressive Zero-Shot performance, ranging from 39.15\% for \texttt{OpenEA d-w} to 83.85\% \texttt{WK3l15k en-de}.
Thus, even in the weakest setting, approximately 40\% of the entities can be aligned solely from their label, without any sophisticated method.
Consequently, this highlights that comparison against methods not using this information is unfair.
For \texttt{DBP15k}, we can compare the initialization from Wu et al.~\cite{DBLP:conf/ijcai/WuLF0Y019}, used, e.g., by \acrshort{rdgcn} to the performance of the initialization by Xu et al.~\cite{DBLP:conf/iclr/FeyL0MK20}, used, e.g., by \acrshort{dgmc}.
We observe that Wu's initialization is 7-9\% points stronger than Xu's initialization.
For \texttt{OpenEA d-w} we obtain 39.15\% zero-shot performance, despite the original labels of the \texttt{w} side being meaningless identifiers.
This is only due to using attribute triples with a pre-defined set of ``name" attributes, cf. Table~\ref{tab:openea:labels}.

\paragraph{Model Performance}
When comparing the performance of both analyzed models, we can observe that they have a clear advantage over both baselines in two of three datasets. 
However, we cannot identify a single winner among them.
Although the performance of \acrshort{dgmc} dropped compared to the results reported originally\footnote{As a general rule, the results improve by 1-2 points when trained on train+validation, and it is not going to change the picture.}, it still leads by about 3-4 points on almost all DBP15k subsets. 
Therefore, it confirms our observation that a smaller test set automatically leads to better results.
Furthermore, we can see that different initialization with entity name also affects model performance, which especially applies to the ja-en subset for \acrshort{dgmc} or fr-en for \acrshort{gcnalign}.
\acrshort{rdgcn} has a clear advantage on the OpenEA subsets extracted from DBPedia with a margin of between 10 and 13 points on both subsets. 
Note that we significantly improved results of \acrshort{rdgcn} on the OpenEA dataset through our extensive hyperparameter search compared to the original evaluation~\cite{DBLP:journals/pvldb/SunZHWCAL20}.
Interestingly, as can be seen in the next section, the main reason is \emph{not} the exploiting of information about different relations. 
The WK3L15k dataset constitutes an interesting exception.
The performance of the \acrshort{dgmc} method, which is supposed to be robust against noise due to its correspondence refinement, is not better than the zero-shot results. 
While \acrshort{rdgcn} and \acrshort{gcnalign} can improve the results, the improvement by 1-2 points does not look very convincing.
From these results, we conclude that there exists no silver bullet for the task of \gls{ea}, and the method itself is still a hyperparameter.
At the same time, we see that the most realistic dataset poses a real challenge for \gls{sota} methods.

\subsection{Ablation: \acrshort{rdgcn}}
\begin{table}[t]
    \centering
    \caption{
    Ablation results for \acrshort{rdgcn} on OpenEA datasets.
    The setting used by \cite{DBLP:conf/ijcai/WuLF0Y019} is underlined.
    The first number is validation H@1, the second number test H@1.
    Bold highlights the best configuration.
    Please notice that due to the specialties of \gls{ea} evaluation, the test and validation performance are \emph{not} directly comparable~\cite{DBLP:journals/corr/abs-2002-06914}.
    }
    \label{tab:res:ablation}
    \tablesize{}
    \begin{tabular*}{\textwidth}{ll*{4}{@{\extracolsep{\fill} }c}}
\toprule
&& \multicolumn{4}{c}{subset} \\
parameter & value &                              d-w &                              d-y &                   en-de &                            en-fr   \\
\midrule normalization & always &  \textbf{84.06} / \textbf{64.28} &                    99.44 / 97.48 &  97.72 / \textbf{93.56} &  \textbf{96.89} / \textbf{91.52} \\
                        & initial &                    82.67 / 62.58 &           \textbf{99.78} / 98.41 &           97.67 / 93.02 &                    95.56 / 89.50 \\
                        & never &                    78.39 / 61.77 &           99.72 / \textbf{98.53} &  \textbf{98.11} / 80.03 &                    95.44 / 90.14 \\
\midrule GCN & 0 &                    57.33 / 50.79 &                    92.33 / 83.83 &  \textbf{98.11} / 80.03 &                    92.22 / 86.94 \\
layers                        & 1 &                    73.33 / 56.66 &                    99.33 / 98.15 &           96.00 / 91.63 &                    94.50 / 90.49 \\
                        & 2 &                    78.39 / 61.77 &                    99.56 / 98.16 &  97.72 / \textbf{93.56} &  \textbf{96.89} / \textbf{91.52} \\
                        & 3 &  \textbf{84.06} / \textbf{64.28} &  \textbf{99.78} / \textbf{98.41} &           97.00 / 92.18 &                    95.44 / 90.14 \\
\midrule interaction  & 0 &                    78.11 / 60.53 &           99.72 / \textbf{98.53} &  97.72 / \textbf{93.56} &                    95.33 / 89.08 \\
layers                        & 1 &                    78.39 / 61.77 &           \textbf{99.78} / 98.41 &           97.67 / 92.59 &                    95.44 / 90.14 \\
                        & 2 &                    82.67 / 62.58 &                    99.56 / 98.16 &  \textbf{98.11} / 80.03 &  \textbf{96.89} / \textbf{91.52} \\
                        & 3 &  \textbf{84.06} / \textbf{64.28} &                    99.50 / 97.85 &           97.67 / 93.02 &                    95.56 / 89.50 \\
\midrule trainable  & no &  \textbf{84.06} / \textbf{64.28} &           99.72 / \textbf{98.53} &  97.72 / \textbf{93.56} &  \textbf{96.89} / \textbf{91.52} \\
embeddings                        & yes &                    82.67 / 62.58 &           \textbf{99.78} / 98.41 &  \textbf{98.11} / 80.03 &                    95.56 / 89.50 \\
\midrule similarity & cos &                    82.67 / 62.58 &                    99.56 / 98.16 &  \textbf{98.11} / 80.03 &                    95.56 / 89.50 \\
                        & dot &                    63.28 / 40.80 &                    91.50 / 79.81 &           85.17 / 78.54 &                    89.94 / 78.17 \\
                        & l1 (inv.) &                    77.89 / 60.78 &                    99.50 / 97.85 &           93.78 / 88.96 &                    94.06 / 88.69 \\
                        & l1 (neg.) &  \textbf{84.06} / \textbf{64.28} &           99.72 / \textbf{98.53} &  97.72 / \textbf{93.56} &  \textbf{96.89} / \textbf{91.52} \\
                        & l2 (inv.) &                    75.28 / 60.20 &                    96.72 / 92.06 &           95.06 / 90.13 &                    94.44 / 89.60 \\
                        & l2 (neg.) &                    72.50 / 51.04 &           \textbf{99.78} / 98.41 &           94.61 / 89.40 &                    94.28 / 87.79 \\
\midrule hard  & no &                    82.67 / 62.58 &  \textbf{99.78} / \textbf{98.41} &  \textbf{98.11} / 80.03 &  \textbf{96.89} / \textbf{91.52} \\
negatives                        & yes &  \textbf{84.06} / \textbf{64.28} &                    99.67 / 98.30 &  97.72 / \textbf{93.56} &                    95.33 / 90.62 \\
\bottomrule
\end{tabular*} \end{table}
We additionally present the results of an ablation study for some model parameters of \acrshort{rdgcn} on the OpenEA datasets in Table~\ref{tab:res:ablation}.
For each presented parameter and each possible value, we fix this one parameter and select the best configuration among all configurations with the chosen parameter setting according to validation H@1.
The cell then shows the validation and test performance of this configuration.
We highlight the best setting on the respective graph pair in bold font.
Note that the test performance numbers also coincide with the performance reported in Table~\ref{tab:res:all} for OpenEA.
We make the following interesting observations:
for all but one graph pair, \emph{always normalizing} the entity representations before passing them into the layers is beneficial.
For \texttt{d-y}, where this is not the case, the difference in performance is small.
For the \emph{number of GCN layers}, we observe an increase in performance from 0 to 2 layers, and on some datasets (\texttt{d-w}, \texttt{d-y}) even beyond.
Thus, aggregating the entities' neighborhood seems beneficial, highlighting the importance of the graph structure.
For the \emph{number of interaction layers}, which perform \emph{relation-aware} message passing, we observe that for two of the four subsets (\texttt{d-y}, \texttt{en-de}) the best configuration does not use any interaction layer.
However, the difference is small.
None of the best configurations uses \emph{trainable node embeddings}.
The \emph{negative $l_1$ similarity} is superior on all datasets, with most of the others being close to it.
Using the dot product seems to be sub-optimal, maybe due to its unbound value range.
Regarding \emph{hard negative mining}, there is no clear tendency, but considering the hard negatives' expensive calculation (all-to-all kNN), its use might not be worthwhile.
Another observation is that sometimes there is a huge gap between the test performance for the best configuration according to validation performance and the best configuration according to test performance.
For instance, if we had selected the hyperparameters according to test performance for \texttt{en-de}, we had obtained 93.53 H@1, while choosing them according to validation performance results in only 80.03 H@1 -- a difference of 13.5\% points.
This difference emphasizes the need for a fair hyperparameter selection.

\section{Conclusion}
\label{sec:conclusion}
In this paper, we investigated \acrlong{sota} in \acrlong{ea}.
Since we identified shortcomings in the commonly employed evaluation procedure, including the lack of validation sets for hyperparameter tuning and different initializations, we provided a fair and sound evaluation over a wide range of configurations. We additionally gave insight into the importance of individual components.
Our results provide a strong, fair, and reproducible baseline for future works to compare against and offer deep insights into the inner workings of a \acrshort{gnn}-based model.

We plan to investigate the identified weakness against noisy labelings in future work and increase the robustness.
Moreover, we aim to improve the usage of relation type information in the message passing phase of models like \acrshort{rdgcn}, which only use them in an initial entity representation refinement stage. For some datasets such as OpenEA d-y and en-de, optimal configurations did not consider the relational information. However, intuitively, this information should help to improve the structural description of entities. Potential improvements include establishing a relation matching between the two graphs or modifying the mechanism used to integrate relational information. 

\section*{Acknowledgements}
This work has been funded by the German Federal Ministry of Education and Research (BMBF) under Grant No. 01IS18036A. The authors of this work take full responsibilities for its content.

\bibliographystyle{plain}
\bibliography{main}

\end{document}